\documentclass[runningheads]{llncs}
\usepackage{graphicx}
\usepackage{array}
\newcolumntype{L}{>{\centering\arraybackslash}m{3cm}}
\usepackage{adjustbox}
\usepackage{graphicx}
\usepackage{amsmath}
\usepackage{wrapfig}
\usepackage{hhline}
\usepackage{hyperref}
\usepackage{subfig}

\begin{document}
\title{Evaluation and Analysis of Different Aggregation and Hyperparameter Selection Methods for Federated Brain Tumor Segmentation}
% Federated Brain Tumor Segmentation on Multi-Institutional Collaborative Learning
% Federated Learning for Brain Tumor Segmentation on Multi-Institutional Datasets

\author{Ece Isik-Polat\inst{1}\orcidID{0000-0002-0728-5390} \and \\
Gorkem Polat\inst{1}\orcidID{0000-0002-1499-3491} \and \\
Altan Kocyigit\inst{1}\orcidID{0000-0001-5003-4127} \and \\
Alptekin Temizel\inst{1,2}\orcidID{0000-0001-6082-2573}}

\authorrunning{Isik-Polat et al.}
% If there are more than two authors, 'et al.' is used.https://www.overleaf.com/project/60eb057a89ab3eb508c13103
%
\institute{Graduate School of Informatics, Middle East Technical University, Ankara, Turkey 
\email{\{eceisik, gorkem.polat, kocyigit, atemizel \}@metu.edu.tr}
\and
Neuroscience and Neurotechnology Center of Excellence, Ankara, Turkey
}
\maketitle    % typeset the header of the contribution
\begin{abstract}

Availability of large, diverse, and multi-national datasets is crucial for the development of effective and clinically applicable AI systems in the medical imaging domain. However, forming a global model by bringing these datasets together at a central location, comes along with various data privacy and ownership problems. To alleviate these problems, several recent studies focus on the federated learning paradigm, a distributed learning approach for decentralized data. Federated learning leverages all the available data without any need for sharing collaborators' data with each other or collecting them on a central server. Studies show that federated learning can provide competitive performance with conventional central training, while having a good generalization capability. In this work, we have investigated several federated learning approaches on the brain tumor segmentation problem. We explore different strategies for faster convergence and better performance which can also work on strong Non-IID cases.

\end{abstract}

\section{Introduction}

Computer-aided approaches utilizing deep learning models have become prominent in the domain of medical image processing \cite{litjens2017survey}. The amount and diversity of training data used to develop these models are important for model success and generalizability \cite{rieke2020future,sheller2020federated,Sun_2017_ICCV}. Currently, the inadequacy of medical data sources and labeled data have become a bottleneck and led to poor performance of the deep learning based solutions \cite{yang2019federatedconceptandapplications}. In order to overcome these issues, there are several initiatives to form diverse datasets to train reliable and robust models that have good generalization ability and clinical usability. EndoCV Challenges incorporates diverse endoscopy video frames from several institutions worldwide, including different modalities and organs to utilize deep learning methods to detect artifacts and diseases \cite{ali2021deep,ali2020objective,polat2021polyp,polat2020endoscopic}. BraTS Challenges brings multi-institutional multi-parametric magnetic resonance imaging (mpMRI) scans for the analysis of brain tumors and the dataset has been continuously growing \cite{bakas2018identifyingBratsData}. Although these initiatives are very important for reliable and clinical-ready models, they are not feasible to scale because it requires a tremendous work. First of all, it is difficult to represent whole distribution (e.g., minority and under-represented groups) as it requires healthy collaborations with many institutions and immense annotations. Secondly, data properties such as image modalities and resolutions are in a constant change that leads to distribution shift over time; therefore, collecting and processing all the data for once does not work either. Moreover, due to the data privacy regulations, collecting sensitive patient data from different institutions and hospitals is not always applicable. The federated learning (FL) concept offers a solution in such situations where data privacy and ownership are a problem by enabling collaborators train a common global model without disclosing their local data \cite{li2020federated,mcmahan2017communication}. Several studies employing FL approaches in the medical domain have reported successful results \cite{dou2021federated,rieke2020future,sheller2020federated}. These studies have drawn the attention of researchers into FL for medical imaging and made it a popular research field recently. 

In this study, we propose various FL approaches for the Federated Tumor Segmentation (FeTS) Challenge \cite{pati2021federated}. For the Task-1 of the challenge, the participants are provided with an FL environment setup that is based on the OpenFL \cite{reina2021openfl} framework and they are requested to develop strategies for the development of the methods in order to extract much of the knowledge from the collaborators. In this task, the participants are allowed to modify four functions: 1) custom performance metrics, 2) collaborator selection, 3) hyperparameter selection, and 4) custom aggregator. Our proposed methods took the 3rd place in the competition.

\section{Related Work}

Recently, the use of FL has been increasing in the medical field. In \cite{huang2020loadaboost}, Huang et al. proposed Loss-based Adaptive Boosting Federated Averaging (LoAdaBoost FedAvg) on critical care database data called as MIMIC-III \cite{johnson2016mimic}. In this method, the collaborators with higher losses than the previous round median loss are retrained before sending to the server for model aggregation. In \cite{li2019privacy}, Li et al. have proposed a federated learning system for brain tumor segmentation on BraTS 2018 dataset \cite{bakas2018identifyingBratsData} and have shown the trade-off between privacy protection costs and model performance. Similarly, in \cite{sheller2020federated}, Sheller et al. have compared federated learning and other data private collaborative learning approaches such as institutional incremental learning and cyclic institutional incremental learning on brain tumor segmentation task. This study has shown that FL can overcome institutional biases and form a global model that has better generalization where data amount and data diversity are inadequate. In \cite{dou2021federated}, Dou et al. have used FL architecture to detect chest CT abnormalities in COVID-19 patients and showed that federated global model outperforms in terms of generalizability on external datasets better than individual models and their ensemble.

\section{Data}

% \begin{wrapfigure}{r}{0.49\textwidth}
% \vspace*{-25mm}
% \includegraphics[trim={2.0cm 1.0cm 0.5cm 1.0cm},clip,width=0.50\textwidth]{images/natural_split_circle.pdf} 
% \vspace*{-8mm}
% \caption{Natural split of the training data}
% \vspace*{-8mm}
% \label{fig:natural_split_training_circle}
% \end{wrapfigure}

\begin{figure}%
    \centering
    \subfloat[\centering Natural split, \textit{partitioning\_1}.]{{\includegraphics[trim={2.0cm 1.0cm 0.1cm 1.0cm},clip,width=0.48\textwidth]{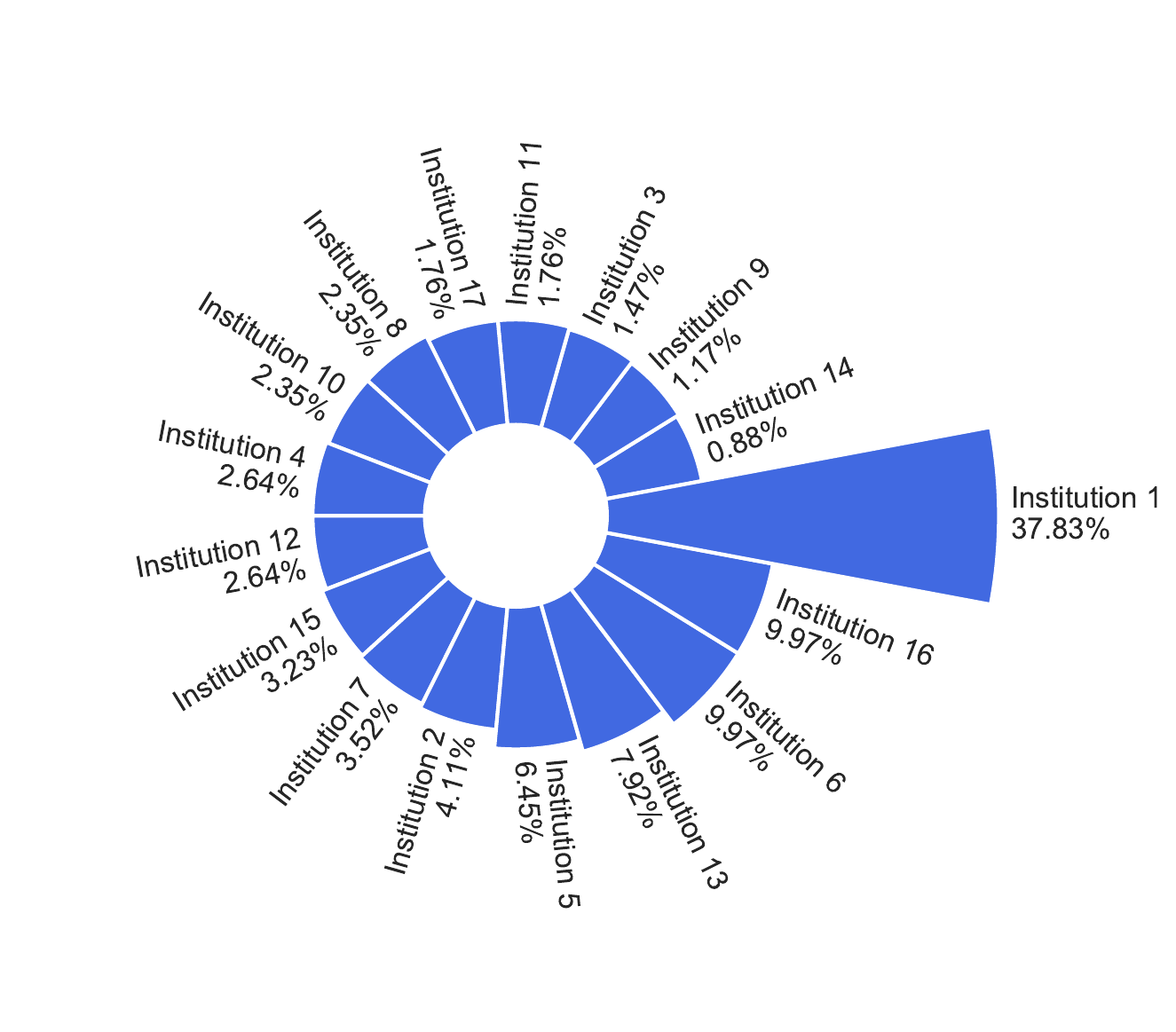} }}%
    \qquad
    \subfloat[\centering Artificial split, \textit{partitioning\_2}:\newline 5 largest institutions are splitted further according to the tumor sizes.]{{\includegraphics[width=0.42\textwidth]{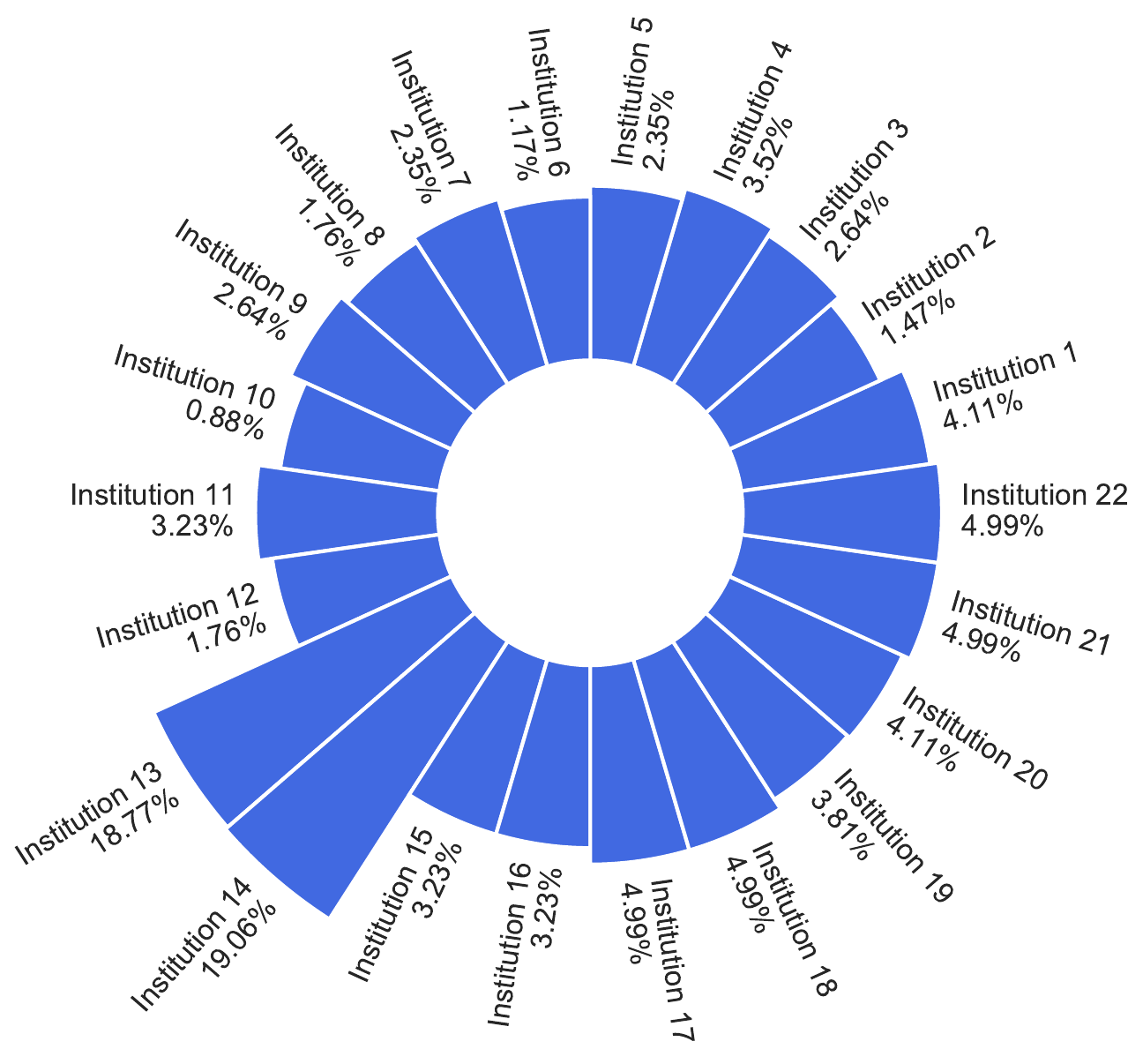} }}%
    \caption{The data distribution in the training dataset splits.}%
    \label{fig:data_split}%
\end{figure}

The Federated Tumor Segmentation (FeTS) challenge 2021 is the first challenge in the federated medical imaging area. The challenge data set is composed of multi-institutional magnetic resonance images from the International Brain Tumor Segmentation (BraTS) challenge and  other independent institutions in the FeTS initiative \cite{bakas2017segmentationgbm,bakas2017segmentationlgg,bakas2017advancing,pati2021federated,reina2021openfl}. The training set contains 341 images, institution-based split of which is given in Fig. \ref{fig:data_split}. The validation and the tests sets contain 111 and 166 images, respectively. The segmentation annotations of the challenge dataset were performed by annotators whose experience levels vary with respect to their clinical and academic backgrounds. Then, these annotations were approved by two experienced board-certified neuroradiologists with more than 12 years of experience \cite{pati2021federated}.

\section{Methods}

\subsection{Aggregator}

In a real-life FL setting, the data distribution of collaborators is non independent identically distributed (non-IID) because collaborators may have different data distribution and the number of observations. The difference in device capabilities, user demographic information, or geographic location can be major reasons for the non-IIDness \cite{kairouz2019advances,mcmahan2017communication}. 

\begin{figure}[ht!]
  \centering
  \includegraphics[width=0.60\textwidth]{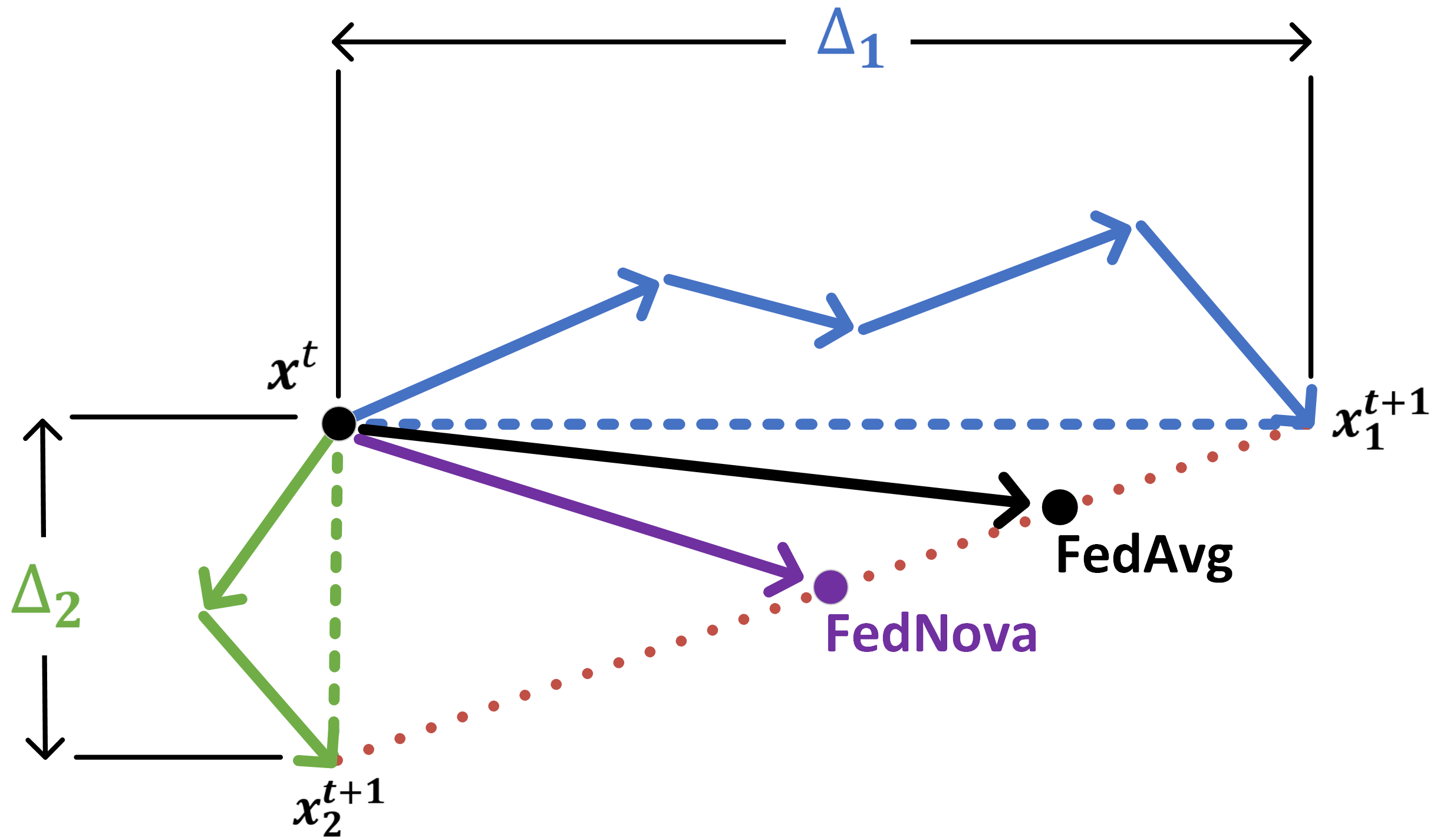}
  \caption{Naive weighted averaging (FedAvg) creates bias toward collaborators having higher number of  samples, which may adversely affect the out-of-distribution performance. On the other hand, FedNova gives equal weights to all collaborators acting as a regularizer.}
  \label{fig:fedNova}
\end{figure}

When collaborators have access to differing amounts of data and when they use the same number of epochs $E$ in their local training, they would perform different numbers of local updates $\tau$. If a collaborator has $n_i$ samples, number of local gradient descent (GD) iterations is $\tau_i = En_i/B$, where $B$ is the mini-batch size. In \cite{wang2020tackling}, Wang et al. have shown that the heterogeneity in collaborators' local progresses causes convergence to a stationary point of mismatched objective function, which is different from the true objective, when vanilla weighted averaging is used. Instead, they propose FedNova, a normalized averaging method that prevents bias toward clients performing more local updates. The shared global model is updated as in Eq. \ref{eqn_fednova}.

\begin{equation} \label{eqn_fednova}
    \boldsymbol{x}^{t+1}=\boldsymbol{x}^t-\tau_\mathrm{eff} \sum\limits_{i=1}^m p_i \frac{\Delta_i^t}{\tau_i^t}
\end{equation}
where $p_i$ denotes the relative sample size of the collaborator $i$ (i.e., $p_i=n_i/n$ where n is the total number of samples), $\tau_\mathrm{eff}=\sum_{i=1}^m p_i \tau_i$, $\Delta_i^t = \boldsymbol{x}^t-\boldsymbol{x}_i^{t+1}$, and $m$ is the total number of collaborators. Since the number of samples $n_i$ for collaborator $i$ is directly proportional to the number of local iteration $\tau_i$ and the relative sample size $p_i$, this formula can be rewritten as in Eq. \ref{eqn_fednova_simplified}.

\begin{gather*} \label{eqn_fednova_reduction}
    \tau_i=\frac{p_i n E}{B} \\
    \tau_\mathrm{eff}=\sum\limits_{i=1}^m \frac{p_i^2 n E}{B} \\
    \boldsymbol{x}^{t+1}=\boldsymbol{x}^t - \frac{nE}{B}(p_1^2+p_2^2+ \cdots +p_m^2) \sum \limits_{i=1}^m \frac{B}{nE} \Delta_i^t \\
    \boldsymbol{x}^{t+1}=\boldsymbol{x}^t-\underbrace{(p_1^2+p_2^2+\cdots +p_m^2)}_{constant} \sum \limits_{i=1}^m \Delta_i^t 
\end{gather*}
\begin{equation} \label{eqn_fednova_simplified}
    \boldsymbol{x}^{t+1}=\boldsymbol{x}^t-\gamma \sum\limits_{i=1}^m \Delta_i^t
\end{equation}
where $\gamma$ refers to the aggregator learning rate, which can be increased or decreased according to FL training needs. As given in Eq. \ref{eqn_fednova_simplified}, FedNova corresponds to a uniform averaging with adjustable step size (or learning rate) on the aggregator. FedNova aims to prevent exacerbation of client drifts caused by relative sample sizes $p_i$. When there is a significant difference between the number of samples in the collaborators, as in the FeTS Challenge dataset, FedAvg creates a bias toward the collaborators having more samples (Fig. \ref{fig:fedNova}). Although validation set (named Val-1 in section \ref{experimental_results}) results reported during the training may seem good as its data distribution directly comes from the training set, out-of-distribution performance results may not be satisfactory. Wang et al. \cite{wang2020tackling} have shown that FedNova generally achieves 6-9\% higher accuracy than FedAvg on a non-IID version of CIFAR-10 dataset.

Another approach to deal with convergence issues when collaborators' data distribution is non-IID is Federated Averaging with server momentum (FedAvgM). The momentum on top of the Stochastic Gradient Descent (SGD) has proven to provide a significant success in accelerating the training and dampening the oscillations \cite{Goodfellow-et-al-2016}. In \cite{hsu2019measuring}, Hsu et al. have shown that as the level of non-IIDness increases, the performance of the FedAvgM stays relatively constant while federated averaging falls rapidly. Moreover, \cite{reddi2021adaptive} has shown the improved effect of adaptive optimizers such as Adam and RMSProp, which are based on momentum, on top of the federated averaging. 

In FedAvgM, the average of the gradients are added to the accumulated gradient which is multiplied by a $\beta$ parameter to adjust effect of the momentum as shown in Eq.~\ref{eqn_fedmomentum}. Then this weighted accumulated gradient is used to update weights of the current communication round as in Eq.~\ref{eqn_fedavgm}. Here, an aggregator learning rate $\gamma$ can be used to adjust the step size on the server (in our experiments $\beta$ is chosen as 0.9 and $\gamma$ is chosen as 1).

\begin{gather*}
    \Delta \boldsymbol{w}^{t+1} = \sum\limits_{i=1}^m p_i \Delta \boldsymbol{w_i}^{t+1}
\end{gather*}
\begin{equation} \label{eqn_fedmomentum}
    \boldsymbol{v}^{t+1}=\beta \boldsymbol{v}^t + \Delta \boldsymbol{w}^{t+1}
\end{equation}
\begin{equation} \label{eqn_fedavgm}
    \boldsymbol{w}^{t+1}=\boldsymbol{w}^t- \gamma \boldsymbol{v}^{t+1}
\end{equation}

where $p_i$ denotes the relative sample size of the collaborator $i$ (i.e., $p_i=n_i/n$ where n is the total number of samples), and $m$ is the total number of collaborators.

Along with FedNova and FedAvgM, other aggregator functions (Table \ref{tab:aggregator_method_list}) have been implemented and experimented in the FeTS Challenge. However, in this article, only the results for FedNova and FedAvgM are presented. Please visit \url{https://github.com/eceisik/FeTS_Challenge_METU_FL_Team} to see all implemented methods by the METU FL Team.

\begin{table}[ht!]
\caption{The list of other aggregator methods implemented.}
\label{tab:aggregator_method_list}
\begin{adjustbox}{width=\columnwidth, center}
\begin{tabular}{|L|L|}\hline
\textbf{Function Name} & \multicolumn{1}{m{8cm}|}{\textbf{Explanation} } \\\hline

make aggregation with improved nodes &   \multicolumn{1}{m{8cm}|}{All collaborators participate in each round of FL, but only those that have improved validation scores participate in the main model aggregation.}\\\hline

 coordinate-wise median aggregation   &   \multicolumn{1}{m{8cm}|}{Main model weights are determined by taking the median of the collaborators weights. Median is more robust to the outliers and extreme values than mean. }\\\hline

\end{tabular}
\end{adjustbox}
\end{table}

\subsection{Collaborator Selection}

% KAIROUZların büyük paperda diyor ki cross siloda tüm collaboratorlar  katılır bunu atıfta bulunarak böyle sectikdiyebiliriz

How to choose collaborators that will take part in each round is another important dimension of the FeTS Challenge. We used "all\_collaborators\_train" as a collaborator choice function and all collaborators participated in each FL round.

We implemented two alternative collaborator choice functions given in Table \ref{tab:collaborator_method_list}. If the focus is on the convergence time metric, the method called as "choose random nodes with faster ones" could be more preferable. This method does not introduce any extra communication delays, because once a random collaborator is selected, only those that are faster than the selected one participate in the training for the FL round (i.e., selected collaborator creates an upper bound for the other selected collaborators in terms of time). Although the number of collaborators participate in each round varies, the working mechanism tends to favor the fastest collaborators. Being fast, in this case, depends on two factors namely the amount of available computation/communication resources and the number of samples in a collaborator. On the other hand, the institutions having fewer patient images may be over represented, which is a disadvantage of this method.

\begin{table}[ht!]
\caption{The list of other collaborator choice methods implemented.}
\label{tab:collaborator_method_list}
\begin{adjustbox}{width=\columnwidth, center}
\begin{tabular}{|L|L|}\hline
\textbf{Function Name} & \multicolumn{1}{m{8cm}|}{\textbf{Explanation} } \\\hline

choose random nodes with faster ones &   \multicolumn{1}{m{8cm}|}{For the first round, all collaborators participate in the training and round time statistics are recorded. After the first round, a random collaborator is selected and participates in the current FL round with other collaborators that are faster than itself. 
 }\\\hline

random collaborators train &   \multicolumn{1}{m{8cm}|}{A random subset of collaborators are selected for each FL round.}\\\hline
\end{tabular}
\end{adjustbox}
\end{table}

\subsection{Hyperparameter Selection}

For the hyperparameter selection, an adaption of AdaComm \cite{MLSYS2019_c8ffe9a5_adacomm} with a learning rate scheduling scheme is used. AdaComm \cite{MLSYS2019_c8ffe9a5_adacomm} is an adaptive communication strategy that saves communication delay and enables fast convergence by federated averaging less frequently in early training rounds and later increasing communication frequency. In \cite{MLSYS2019_c8ffe9a5_adacomm}, experimental converge analysis was examined on wall-clock time instead of communication round. It is shown that using more local updates in the early rounds of training resulted in a faster decrease in loss but also a higher error. For this reason, it starts with a large number of updates per round and gradually decreases as the model starts to converge.

In the original version of AdaComm, the method is based on the number of local updates in an IID setting. However, in the challenge, the data distribution is extremely uneven. While Institute-1 has 37.83\% of the data, Institute-14 has 0.88\% of the whole training data (Figure~\ref{fig:data_split}). Using the same number of local updates for each collaborator could potentially cause over-representation of some small data provider institutions. By considering the non-IID nature of the data distribution, our aggregation method mechanism, and the fact that the number of local updates is directly proportional to the number of epochs, we adapted this method based on the decaying number of epoch (AdaptiveEpoch). Basically, the number of epochs per round at each FL round decays according to the relative difference between the initial loss and current round loss as stated in Eq. \ref{adacom}. \\

\begin{equation} \label{adacom}
    E_t = \Bigg\lceil \sqrt{\frac{F(x_{T=t})}{F(x_{T=0})}}E_0 \Bigg\rceil
\end{equation}
where $T$ denotes the number of FL rounds, $t$ denotes the round number, $E_t$ denotes the number of epochs at a given round $t$, and $F(x)$ is the objective function with respect to model parameters denoted by $x$.

Learning rate scheduling is a commonly used technique to train deep neural networks in a centralized manner \cite{Goodfellow-et-al-2016}. Studies show that learning rate scheduling is also necessary for FedAvg to converge to an optimum point of loss function \cite{li2019convergence}. However, there are many strategies for scheduling and there is no benchmark for their performances. In this study, we have adopted \textit{decay learning rate on plateau} approach. This strategy brings two new parameters namely \textit{patience} and \textit{decay factor}. In our implementation, learning rate scheduling tracks the target performance metric, which is the mean Dice score for ET, TC and WT labels, and if there is no improvement on the target performance metric for a \textit{patience} number of round, the learning rate is updated by scaling with the \textit{decay factor}. Experiments show that learning rate scheduling provides faster convergence, more relaxed learning rate selection, higher convergence score, and reduced oscillations when training converges \cite{Goodfellow-et-al-2016}. The list of hyperparameter selection methods are given in Table~\ref{tab:hyper_method_list}. For AdaptiveEpoch initial epoch $E_0$ is set to 8; for the LR scheduling, the initial LR is set to $0.0002$ and patience is set as 15. For the constant hyperparameters, default values were used (LR=0.00005, epoch per round=1).

\begin{table}[ht!]
\caption{The list of other hyperparameter selection methods.}
\label{tab:hyper_method_list}
\begin{adjustbox}{width=\columnwidth, center}
\begin{tabular}{|L|L|}\hline
\textbf{Function Name} & \multicolumn{1}{m{8cm}|}{\textbf{Explanation} } \\\hline
constant hyperparameters &  \multicolumn{1}{m{8cm}|}{Use fixed hyperparameters for each FL round.} \\\hline
LRScheduling hyperparameters &   \multicolumn{1}{m{8cm}|}{Learning rate decays according to the value of average Dice score with given patience scheme.}\\\hline

AdaptiveEpoch &   \multicolumn{1}{m{8cm}|}{The number of epochs per round decays according to the decrease in the initial loss.}\\\hline

\end{tabular}
\end{adjustbox}
\end{table}

\section{Experimental Results \& Discussion} \label{experimental_results}

Before the FL training, the training dataset is split into train and validation sets as 80\% and 20\%, respectively. The performance results of the aggregated and individual models on validation sets are logged at each FL round (it is integrated with the FeTS Challenge source code). Unless otherwise stated, all reported performance metrics and loss graphs belong to this validation set of \textit{partitioning\_2.csv}. The mean Dice score refers to the average of Dice scores of ET, TC, and WT labels. 

\begin{figure}[htbp!]
  \centering
  \includegraphics[width=\textwidth]{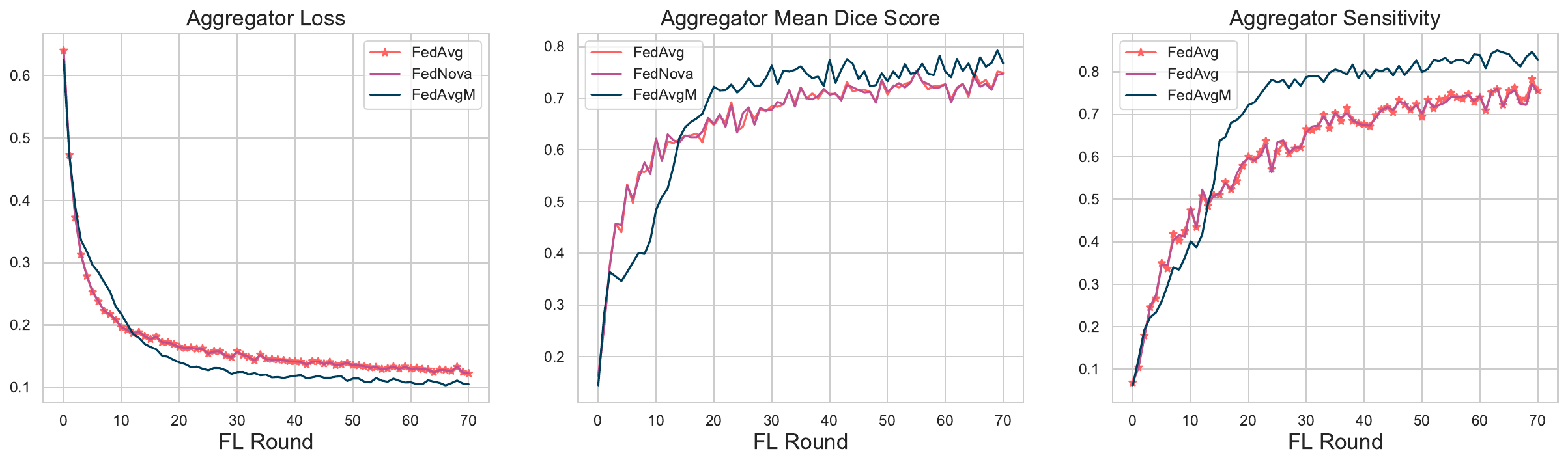}
  \caption{The performance comparison of FedAvg, FedNova, and FedAvgM.}
  \label{fig:fedavg_fednova_fedavgm}
\end{figure}

Figure~\ref{fig:fedavg_fednova_fedavgm} shows the performances of FedAvg, FedNova, and FedAvgM on aggregator mean dice score, aggregator loss, and aggregator sensitivity metrics. Since medical datasets may contain institutional biases \cite{sheller2020federated} and FedAvg have an undesirable effect of favoring these biases, FedNova is expected to have better performance on the non-training sets. However, since samples of institutions' distributions of training and validation sets are similar to each other, we observe nearly identical performance for both FedAvg and FedNova. Yet, models built with FedNova are expected to have better inferences on the out-of-distribution dataset \cite{wang2020tackling}, and as such, they are expected to be more suitable for real-life use-case scenarios. On the other hand, FedAvgM outperforms both FedAvg and FedNova on all metrics. Therefore, we have preferred FedAvgM as the aggregator method in the FeTS challenge.

\begin{figure}[ht!]
% \vspace{-3mm}
  \centering
  \includegraphics[width=\textwidth]{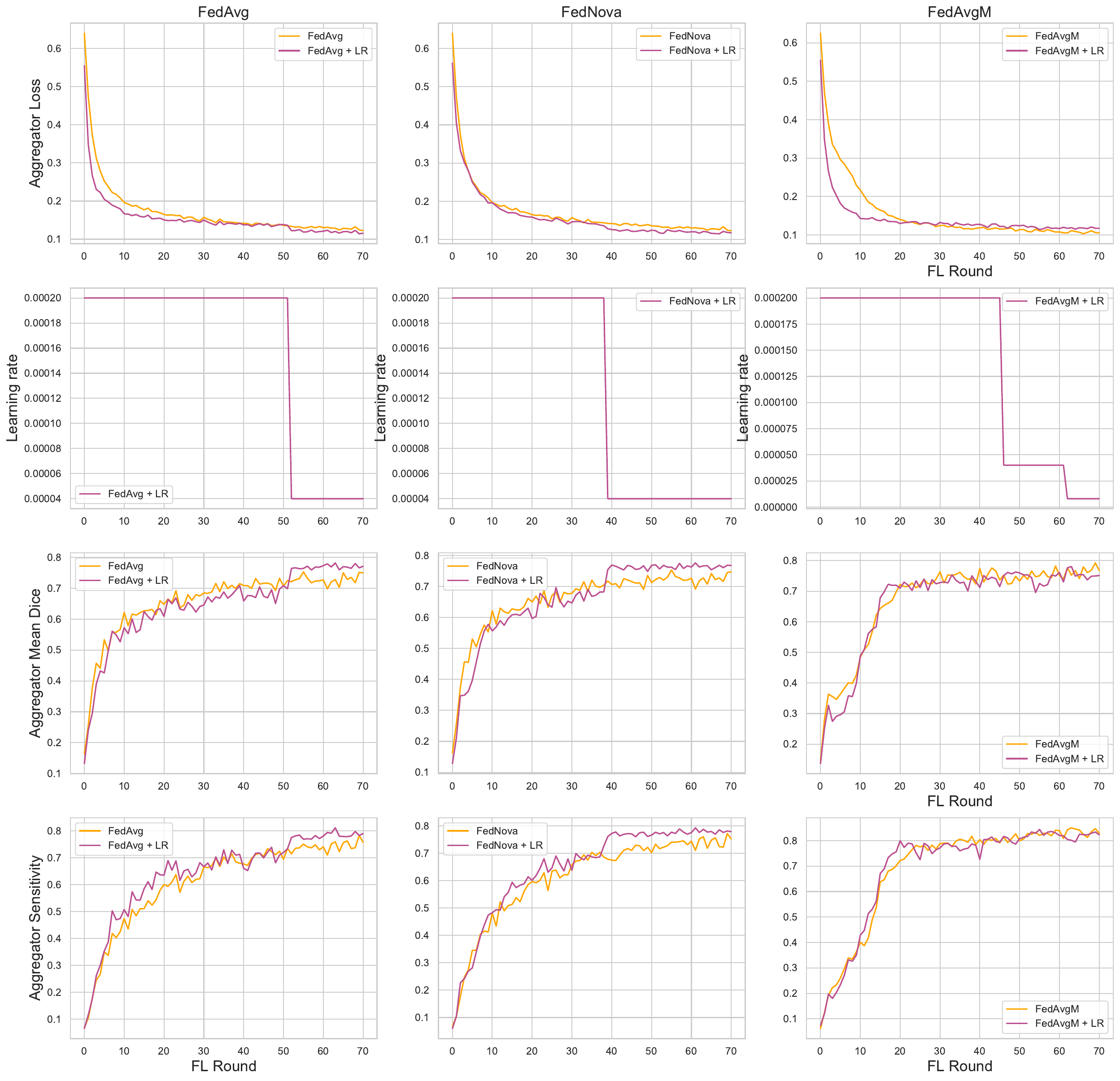}
  \caption{The impacts of LR scheduling on FedAvg, FedNova, and FedAvgM.}
  \label{fig:lr_effect}
\end{figure}

Figure~\ref{fig:lr_effect} shows the effect of the LR scheduling approach on each aggregation method. For both FedAvg and FedNova, it can be observed on both loss function and performance metrics that LR scheduling has an evident effect on their performances. In particular, a sharp increase on performance metrics occurs when the LR is decayed. On the other hand, LR scheduling has no improvement on FedAvgM. One possible reason might be since FedAvgM converges much faster than the FedAvg and FedNova, it may have directly reached the optimum region where it does not need any scheduling. However, it should be noted that we have used fixed values for starting learning rate, decay rate, and patience parameters; therefore, more experiments with different set of values should be performed to make a comment on effect of LR scheduling on FedAvgM.

\begin{figure}[h!]
  \centering
  \includegraphics[width=\textwidth]{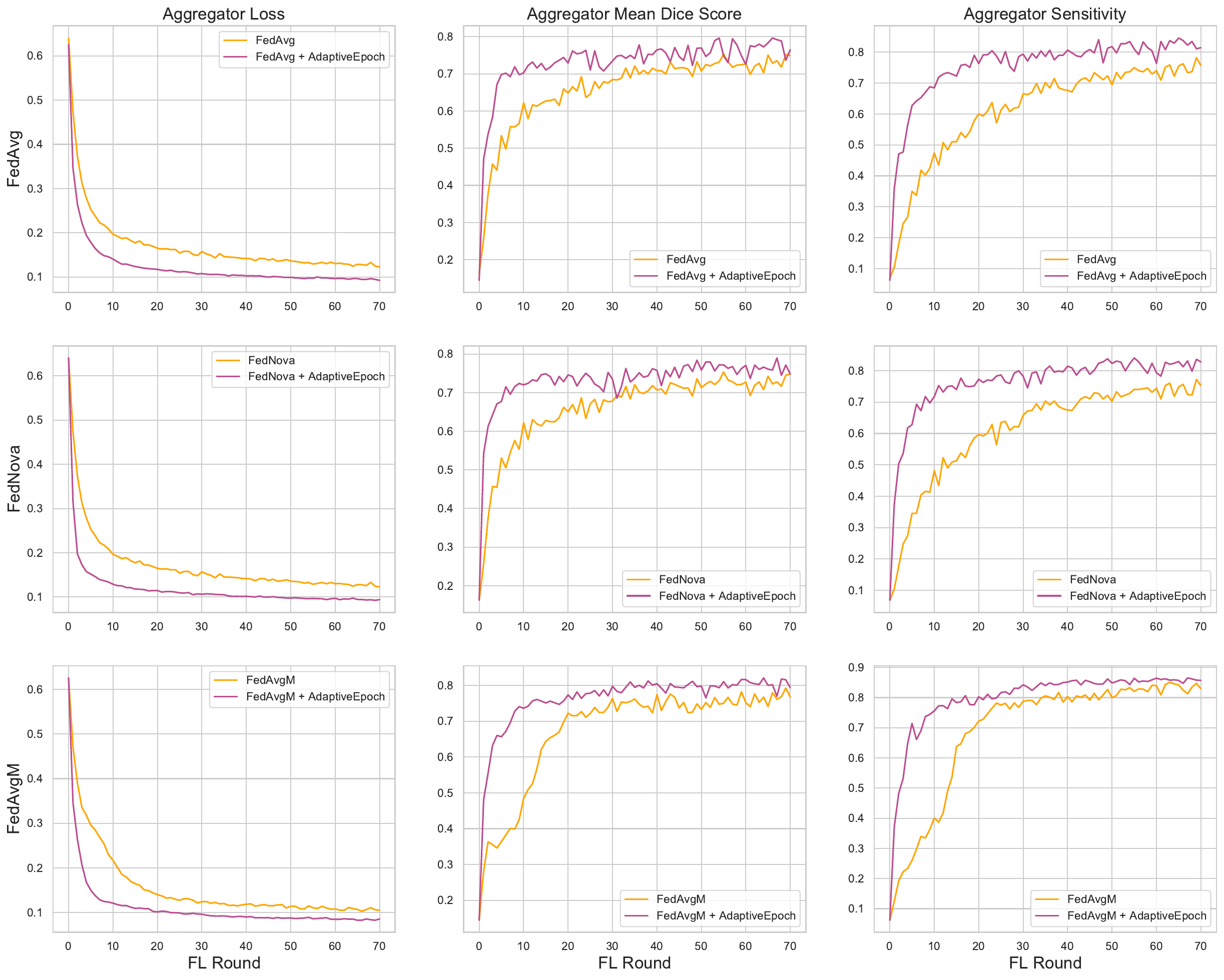}
  \caption{The impacts of adaptive epoch on FedAvg, FedNova, and FedAvgM. }
  \label{fig:adaptive_epoch_effect}
\end{figure}

AdaptiveEpoch helps training converge in fewer rounds with higher performance due to having more local epochs than using the constant hyperparameters as seen in Figure~\ref{fig:adaptive_epoch_effect}. The AdaptiveEpoch method improves the performance of both FedAvg, FedNova, and FedAvgM on aggregator loss, aggregator mean dice score, and aggregator sensitivity metrics.  The improvement achieved on aggregator methods by AdaptiveEpoch is much more significant than the LR scheduling. The performance increase can be observed both on loss and performance metrics.

\begin{figure}[htbp!]
  \centering
  \includegraphics[width=\textwidth]{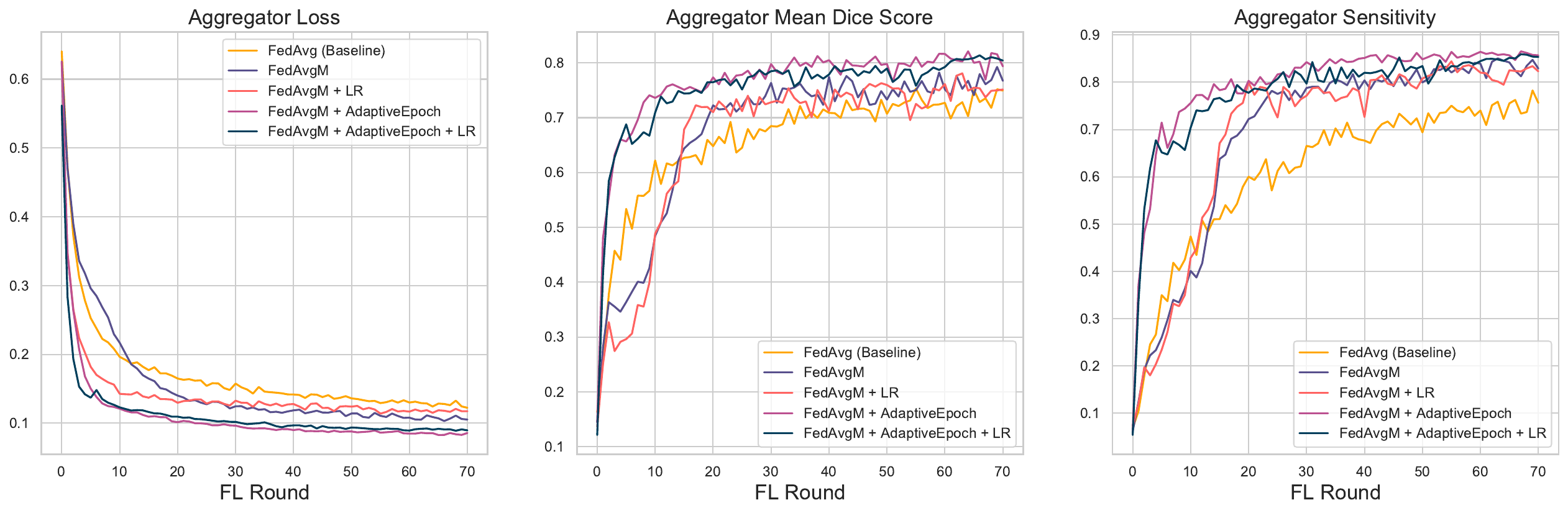}
  \caption{The impacts of hyperparameter setting strategies on FedAvgM }
  \label{fig:fedavgm_all_combinations}
\end{figure}

Figure~\ref{fig:fedavgm_all_combinations} shows the performance comparison of different hyperparameter strategies on FedAvgM. Accordingly, LR scheduling, AdaptiveEpoch, and AdaptiveEpoch+LR scheduling improves the baseline model performance. AdaptiveEpoch and AdaptiveEpoch+LR scheduling provides faster convergence than LR scheduling. However, there is no significant difference between AdaptiveEpoch and AdaptiveEpoch+LR scheduling. Due to the time and resource constraints, the number of FL round was set to 70 for all experiments, which in turn limited the effect of LR scheduling and AdaptiveEpoch+LR scheduling due to incomplete decaying of LR.

Table~\ref{tab:experiment_results} shows the mean dice score and convergence score obtained on the validation set. These experiments are performed by using \textit{partitioning-2} as data split. The convergence score is computed as the area under the validation learning curve where the horizontal axis is the runtime, and the vertical axis is the performance. Most of the time, FedAvgM outperforms others and achieves the best mean Dice score and the convergence score for all hyperparameter choice strategies except for LR scheduling. It is expected and in line with the results that are presented in Figure~\ref{fig:lr_effect}. Nevertheless, the convergence score is based on the validation set reported during the FL training; therefore, the comparison of convergence scores on an out-of-distribution set is still an open question.

\begin{table}[htb!]
\caption{ The mean Dice score and convergence scores on the validation set.}
\label{tab:experiment_results}
\centering
\begin{adjustbox}{width=\columnwidth,center}
\begin{tabular}{lccccccc}
                                      & \multicolumn{3}{c}{\textbf{Validation Mean Dice}}     &  & \multicolumn{3}{c}{\textbf{Convergence Score}}         \\ 
\hhline{~===~===}
                                      & \textbf{FedAvg} & \textbf{FedNova} & \textbf{FedAvgM} &  & \textbf{FedAvg} & \textbf{FedNova} & \textbf{FedAvgM}  \\ 
\hline
\textbf{Constant Hyperparameters}     & 0.753           & 0.753            & \textbf{0.793}   &  & 0.770           & 0.770            & \textbf{0.788}    \\
\textbf{LR Scheduling}                & \textbf{0.782}  & 0.776            & 0.781            &  & \textbf{0.761}  & 0.759            & 0.744             \\
\textbf{AdaptiveEpoch}                & 0.796           & 0.790            & \textbf{0.821}   &  & 0.797           & 0.797            & \textbf{0.804}    \\
\textbf{AdaptiveEpoch+LR Scheduling~} & 0.783           & 0.787            & \textbf{0.814}   &  & 0.781           & 0.785            & \textbf{0.795}    \\
\hline
\end{tabular}
\end{adjustbox}
\end{table}

\clearpage  %fix me
Table~\ref{tab: challenge_combination_results} presents the results of the our challenge submission on the challenge test set with convergence score of 0.770. The results are provided by the FeTS initiative.

\begin{table}[ht!]
\caption{The scores were obtained on Leader Board 2 of Task 1 that our team (METU FL) won the $3^{rd}$ rank in the FeTS Challenge.}
\label{tab: challenge_combination_results}
\centering
\begin{tabular}{lccccc}
                           & \textbf{$\mu$} & \textbf{$\sigma$} & \textbf{$Q_1$} & \textbf{$Q_2$} & \textbf{$Q_3$}  \\ 
\hline\hline
\textbf{Dice\_ET}          & 0.719       & 0.268         & 0.676             &  0.811         & 0.887              \\
\textbf{Dice\_WT}          & 0.794       & 0.228         & 0.807             &  0.882         & 0.917              \\
\textbf{Dice\_TC}          & 0.741       & 0.286         & 0.674             &  0.870         & 0.931              \\ 
\hline
\textbf{Sensitivity\_ET}   & 0.817       & 0.277         & 0.812             &  0.930         & 0.974              \\
\textbf{Sensitivity\_WT}   & 0.860       & 0.236         & 0.869             &  0.942         & 0.977              \\
\textbf{Sensitivity\_TC}   & 0.832       & 0.278         & 0.861             &  0.945         & 0.981              \\ 
\hline
\textbf{Specificity\_ET}   & 0.999       & 0.002         & 0.999             &  0.999         & 0.999              \\
\textbf{Specificity\_WT}   & 0.998       & 0.004         & 0.998             &  0.998         & 0.999              \\
\textbf{Specificity\_TC}   & 0.999       & 0.003         & 0.999             &  0.999         & 0.999              \\ 
\hline
\textbf{Hausdorff95\_ET}   & 34.430      & 96.022        & 1.414             &  2.449         & 9.027              \\
\textbf{Hausdorff95\_WT}   & 19.199      & 57.799        & 3.162             &  5.431         & 10.355             \\
\textbf{Hausdorff95\_TC}   & 35.288      & 91.651        & 2.236             &  6.240         & 17.051             \\ 
\hline
\end{tabular}
\end{table}

\section{Conclusion}

In this study, we perform comprehensive experiments to compare different hyperparameter selection strategies and aggregation methods. The experiments reveal that FedAvgM has better performance than FedAvg and FedNova. Moreover, it is shown that the AdaptiveEpoch approach provides performance increase and faster convergence. However, LR scheduling is not effective with FedAvgM or AdaptiveEpoch. Therefore, it can be said that methods that work well individually may not work well together when combined, or one can reduce the effectiveness of the other. For instance, while AdaptiveEpoch results in better validation mean dice scores and convergence scores than using constant hyperparameter strategy, when it is combined with LR scheduling, all mean dice and convergence scores get worse for all aggregation methods (see Table~\ref{tab:experiment_results}).

During the experiments,  all collaborators have participated in the local training process for all rounds. Instead, collaborator choosing methods such as clustering collaborators based on the update similarity or increasing the likelihood of being chosen collaborators that improved the performance for the random collaborator choice can be utilized to improve performance.

Moreover, in the medical image domain, there is generally high interobserver variability in annotations, which can be considered label noise. For example, if an institution’s label quality is low, the model coming from that institution will adversely affect the global model; therefore, weights coming from that institution should be handled carefully. There are defense mechanisms such as KRUM \cite{blanchard2017machine}, BARFED \cite{isikpolat2021barfed}, or trimmed mean \cite{yin2018byzantine} that can overcome the attacks in federated learning to some extent. These defense strategies may be used to overcome the label noise.

\section{GPU Training Times}

Computation time and cost, as well as energy consumption, are important factors determining the direction of future research and adoption of the technology in real life. Table~\ref{tab:gpu_run_time} shows the detailed GPU training times of the experiments that are run on single NVIDIA A100-80GB GPU. LR scheduling has no significant effect on the training times. On the other hand, although AdaptiveEpoch strategy brings an increase in performance metrics, its usage nearly doubles the total training time due to longer round times. 

\begin{table}[htb!]
\caption{The detailed GPU training times (hour).}
\label{tab:gpu_run_time}
\centering
\begin{adjustbox}{width=\columnwidth,center}
\begin{tabular}{lcccc}
~                                    & \textbf{FedAvg} & \textbf{FedNova} & \textbf{FedAvgM} & \textbf{Total}  \\ 
\hline
\textbf{Constant Hyperparameters}    & 36.7            & 36.4             & 34.8             & \textbf{107.9}  \\
\textbf{LR Scheduling}               & 35.1            & 36.7             & 36.8             & \textbf{108.6}  \\
\textbf{AdaptiveEpoch}               & 67.2            & 69.0             & 68.9             & \textbf{205.1}  \\
\textbf{AdaptiveEpoch+LR Scheduling} & 65.5            & 65.8             & 66.4             & \textbf{197.7}  \\ 
\hline\hline
\textbf{Total}                       & \textbf{204.5}  & \textbf{207.9}   & \textbf{206.9}   & \textbf{619.3}  \\
\hline
\end{tabular}
\end{adjustbox}
\end{table}

\section{Acknowledgments}

This work has been supported by Middle East Technical University Scientific Research Projects Coordination Unit under grant number GAP-704-2020-10071. The numerical calculations reported in this paper were performed using TUBITAK ULAKBIM, High Performance and Grid Computing Center (TRUBA resources).

\clearpage

\bibliographystyle{splncs04}
\bibliography{references}

\begin{thebibliography}{10}
\providecommand{\url}[1]{\texttt{#1}}
\providecommand{\urlprefix}{URL }
\providecommand{\doi}[1]{https://doi.org/#1}

\bibitem{ali2021deep}
Ali, S., Dmitrieva, M., Ghatwary, N., Bano, S., Polat, G., Temizel, A.,
  Krenzer, A., Hekalo, A., Guo, Y.B., Matuszewski, B., et~al.: Deep learning
  for detection and segmentation of artefact and disease instances in
  gastrointestinal endoscopy. Medical image analysis  \textbf{70},  102002
  (2021)

\bibitem{ali2020objective}
Ali, S., Zhou, F., Braden, B., Bailey, A., Yang, S., Cheng, G., Zhang, P., Li,
  X., Kayser, M., Soberanis-Mukul, R.D., et~al.: An objective comparison of
  detection and segmentation algorithms for artefacts in clinical endoscopy.
  Scientific reports  \textbf{10}(1),  1--15 (2020)

\bibitem{bakas2017segmentationgbm}
Bakas, S., Akbari, H., Sotiras, A., Bilello, M., Rozycki, M., Kirby, J.,
  Freymann, J., Farahani, K., Davatzikos, C.: Segmentation labels and radiomic
  features for the pre-operative scans of the tcga-gbm collection. the cancer
  imaging archive. Nat Sci Data  \textbf{4},  170117 (2017)

\bibitem{bakas2017segmentationlgg}
Bakas, S., Akbari, H., Sotiras, A., Bilello, M., Rozycki, M., Kirby, J.,
  Freymann, J., Farahani, K., Davatzikos, C.: Segmentation labels and radiomic
  features for the pre-operative scans of the tcga-lgg collection. The cancer
  imaging archive  \textbf{286} (2017)

\bibitem{bakas2017advancing}
Bakas, S., Akbari, H., Sotiras, A., Bilello, M., Rozycki, M., Kirby, J.S.,
  Freymann, J.B., Farahani, K., Davatzikos, C.: Advancing the cancer genome
  atlas glioma mri collections with expert segmentation labels and radiomic
  features. Scientific data  \textbf{4}(1),  1--13 (2017)

\bibitem{bakas2018identifyingBratsData}
Bakas, S., Reyes, M., Jakab, A., Bauer, S., Rempfler, M., Crimi, A., Shinohara,
  R.T., Berger, C., Ha, S.M., Rozycki, M., et~al.: Identifying the best machine
  learning algorithms for brain tumor segmentation, progression assessment, and
  overall survival prediction in the brats challenge. arXiv preprint
  arXiv:1811.02629  (2018)

\bibitem{blanchard2017machine}
Blanchard, P., El~Mhamdi, E.M., Guerraoui, R., Stainer, J.: Machine learning
  with adversaries: Byzantine tolerant gradient descent. In: Proceedings of the
  31st International Conference on Neural Information Processing Systems. pp.
  118--128 (2017)

\bibitem{dou2021federated}
Dou, Q., So, T.Y., Jiang, M., Liu, Q., Vardhanabhuti, V., Kaissis, G., Li, Z.,
  Si, W., Lee, H.H., Yu, K., et~al.: Federated deep learning for detecting
  covid-19 lung abnormalities in ct: a privacy-preserving multinational
  validation study. NPJ digital medicine  \textbf{4}(1),  1--11 (2021)

\bibitem{Goodfellow-et-al-2016}
Goodfellow, I., Bengio, Y., Courville, A.: Deep Learning. MIT Press (2016),
  \url{http://www.deeplearningbook.org}

\bibitem{hsu2019measuring}
Hsu, T.M.H., Qi, H., Brown, M.: Measuring the effects of non-identical data
  distribution for federated visual classification. arXiv preprint
  arXiv:1909.06335  (2019)

\bibitem{huang2020loadaboost}
Huang, L., Yin, Y., Fu, Z., Zhang, S., Deng, H., Liu, D.: Loadaboost:
  Loss-based adaboost federated machine learning with reduced computational
  complexity on iid and non-iid intensive care data. Plos one  \textbf{15}(4),
  e0230706 (2020)

\bibitem{isikpolat2021barfed}
Isik-Polat, E., Polat, G., Kocyigit, A.: Barfed: Byzantine attack-resistant
  federated averaging based on outlier elimination. arXiv preprint
  arXiv:2111.04550  (2021)

\bibitem{johnson2016mimic}
Johnson, A.E., Pollard, T.J., Shen, L., Li-Wei, H.L., Feng, M., Ghassemi, M.,
  Moody, B., Szolovits, P., Celi, L.A., Mark, R.G.: Mimic-iii, a freely
  accessible critical care database. Scientific data  \textbf{3}(1), ~1--9
  (2016)

\bibitem{kairouz2019advances}
Kairouz, P., McMahan, H.B., Avent, B., Bellet, A., Bennis, M., Bhagoji, A.N.,
  Bonawitz, K., Charles, Z., Cormode, G., Cummings, R., et~al.: Advances and
  open problems in federated learning. arXiv preprint arXiv:1912.04977  (2019)

\bibitem{li2020federated}
Li, T., Sahu, A.K., Talwalkar, A., Smith, V.: Federated learning: Challenges,
  methods, and future directions. IEEE Signal Processing Magazine
  \textbf{37}(3),  50--60 (2020)

\bibitem{li2019privacy}
Li, W., Milletar{\`\i}, F., Xu, D., Rieke, N., Hancox, J., Zhu, W., Baust, M.,
  Cheng, Y., Ourselin, S., Cardoso, M.J., et~al.: Privacy-preserving federated
  brain tumour segmentation. In: International workshop on machine learning in
  medical imaging. pp. 133--141. Springer (2019)

\bibitem{li2019convergence}
Li, X., Huang, K., Yang, W., Wang, S., Zhang, Z.: On the convergence of fedavg
  on non-iid data. arXiv preprint arXiv:1907.02189  (2019)

\bibitem{litjens2017survey}
Litjens, G., Kooi, T., Bejnordi, B.E., Setio, A.A.A., Ciompi, F., Ghafoorian,
  M., Van Der~Laak, J.A., Van~Ginneken, B., S{\'a}nchez, C.I.: A survey on deep
  learning in medical image analysis. Medical image analysis  \textbf{42},
  60--88 (2017)

\bibitem{mcmahan2017communication}
McMahan, B., Moore, E., Ramage, D., Hampson, S., y~Arcas, B.A.:
  Communication-efficient learning of deep networks from decentralized data.
  In: Artificial Intelligence and Statistics. pp. 1273--1282. PMLR (2017)

\bibitem{pati2021federated}
Pati, S., Baid, U., Zenk, M., Edwards, B., Sheller, M., Reina, G.A., Foley, P.,
  Gruzdev, A., Martin, J., Albarqouni, S., Chen, Y., Shinohara, R.T., Reinke,
  A., Zimmerer, D., Freymann, J.B., Kirby, J.S., Davatzikos, C., Colen, R.R.,
  Kotrotsou, A., Marcus, D., Milchenko, M., Nazeri, A., Fathallah-Shaykh, H.,
  Wiest, R., Jakab, A., Weber, M.A., Mahajan, A., Maier-Hein, L., Kleesiek, J.,
  Menze, B., Maier-Hein, K., Bakas, S.: The federated tumor segmentation (fets)
  challenge (2021)

\bibitem{polat2021polyp}
Polat, G., Isik~Polat, E., Kayabay, K., Temizel, A.: Polyp detection in
  colonoscopy images using deep learning and bootstrap aggregation. In:
  Proceedings of the 3rd International Workshop and Challenge on Computer
  Vision in Endoscopy (EndoCV 2021) @ ISBI. vol.~2886, pp. 90--100 (2021)

\bibitem{polat2020endoscopic}
Polat, G., Sen, D., Inci, A., Temizel, A.: Endoscopic artefact detection with
  ensemble of deep neural networks and false positive elimination. In: EndoCV@
  ISBI. pp. 8--12 (2020)

\bibitem{reddi2021adaptive}
Reddi, S.J., Charles, Z., Zaheer, M., Garrett, Z., Rush, K., Kone{\v{c}}n{\'y},
  J., Kumar, S., McMahan, H.B.: Adaptive federated optimization. In:
  International Conference on Learning Representations (2021),
  \url{https://openreview.net/forum?id=LkFG3lB13U5}

\bibitem{reina2021openfl}
Reina, G.A., Gruzdev, A., Foley, P., Perepelkina, O., Sharma, M., Davidyuk, I.,
  Trushkin, I., Radionov, M., Mokrov, A., Agapov, D., et~al.: Openfl: An
  open-source framework for federated learning. arXiv preprint arXiv:2105.06413
   (2021)

\bibitem{rieke2020future}
Rieke, N., Hancox, J., Li, W., Milletari, F., Roth, H.R., Albarqouni, S.,
  Bakas, S., Galtier, M.N., Landman, B.A., Maier-Hein, K., et~al.: The future
  of digital health with federated learning. NPJ digital medicine
  \textbf{3}(1), ~1--7 (2020)

\bibitem{sheller2020federated}
Sheller, M.J., Edwards, B., Reina, G.A., Martin, J., Pati, S., Kotrotsou, A.,
  Milchenko, M., Xu, W., Marcus, D., Colen, R.R., et~al.: Federated learning in
  medicine: facilitating multi-institutional collaborations without sharing
  patient data. Scientific reports  \textbf{10}(1),  1--12 (2020)

\bibitem{Sun_2017_ICCV}
Sun, C., Shrivastava, A., Singh, S., Gupta, A.: Revisiting unreasonable
  effectiveness of data in deep learning era. In: Proceedings of the IEEE
  International Conference on Computer Vision (ICCV) (Oct 2017)

\bibitem{MLSYS2019_c8ffe9a5_adacomm}
Wang, J., Joshi, G.: Adaptive communication strategies to achieve the best
  error-runtime trade-off in local-update sgd. In: Talwalkar, A., Smith, V.,
  Zaharia, M. (eds.) Proceedings of Machine Learning and Systems. vol.~1, pp.
  212--229 (2019),
  \url{https://proceedings.mlsys.org/paper/2019/file/c8ffe9a587b126f152ed3d89a146b445-Paper.pdf}

\bibitem{wang2020tackling}
Wang, J., Liu, Q., Liang, H., Joshi, G., Poor, H.V.: Tackling the objective
  inconsistency problem in heterogeneous federated optimization. Advances in
  Neural Information Processing Systems  \textbf{33} (2020)

\bibitem{yang2019federatedconceptandapplications}
Yang, Q., Liu, Y., Chen, T., Tong, Y.: Federated machine learning: Concept and
  applications. ACM Transactions on Intelligent Systems and Technology (TIST)
  \textbf{10}(2),  1--19 (2019)

\bibitem{yin2018byzantine}
Yin, D., Chen, Y., Kannan, R., Bartlett, P.: Byzantine-robust distributed
  learning: Towards optimal statistical rates. In: International Conference on
  Machine Learning. pp. 5650--5659. PMLR (2018)

\end{thebibliography}

\end{document}